\documentclass[sigconf]{aamas}

\usepackage{balance} 
\usepackage{graphicx} 
\usepackage{amsmath}
\usepackage{algorithm}
\usepackage{algpseudocode}
\usepackage{xcolor, colortbl} 
\usepackage{subcaption} 
\definecolor{yellow}{RGB}{238, 218, 172}
\definecolor{lightgreen}{RGB}{194, 218, 194}

\doi{AGOG8131}

\makeatletter
\gdef\@copyrightpermission{
  \begin{minipage}{0.2\columnwidth}
   \href{https://creativecommons.org/licenses/by/4.0/}{\includegraphics[width=0.90\textwidth]{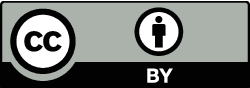}}
  \end{minipage}\hfill
  \begin{minipage}{0.8\columnwidth}
   \href{https://creativecommons.org/licenses/by/4.0/}{This work is licensed under a Creative Commons Attribution International 4.0 License.}
  \end{minipage}
  \vspace{5pt}
}
\makeatother

\setcopyright{ifaamas}
\acmConference[AAMAS '26]{Proc.\@ of the 25th International Conference
on Autonomous Agents and Multiagent Systems (AAMAS 2026)}{May 25 -- 29, 2026}
{Paphos, Cyprus}{C.~Amato, L.~Dennis, V.~Mascardi, J.~Thangarajah (eds.)}
\copyrightyear{2026}
\acmYear{2026}
\acmDOI{}
\acmPrice{}
\acmISBN{}

\title[AAMAS-2026 Formatting Instructions]{IntRec: Intent-based Retrieval with Contrastive Refinement}

\author{Pourya Shamsolmoali}
\affiliation{
  \institution{Dept. Computer Science \\ University of York}
  \country{York, United Kingdom}}

\author{Masoumeh Zareapoor}
\affiliation{
  \institution{Dept. Computer Science \\ Shanghai Jiao Tong University}
  \country{Shanghai, China}}

\author{Eric Granger}
\affiliation{
  \institution{LIVIA, Dept. of Systems Engineering \\ École de technologie supérieure}
  \country{Montreal, Canada}}

\author{Yue Lu}
\affiliation{
  \institution{School of Communication and Electronic Eng. \\ East China Normal University}
  \country{Shanghai, China}}

\begin{abstract}
Retrieving user-specified objects from complex scenes remains a challenging task, especially when queries are ambiguous or involve multiple similar objects. Existing open-vocabulary detectors operate in a one-shot manner, lacking the ability to refine predictions based on user feedback. To address this, we propose IntRec, an interactive object retrieval framework that refines predictions based on user feedback. At its core is an Intent State (IS) that maintains dual memory sets for positive anchors (confirmed cues) and negative constraints (rejected hypotheses). A contrastive alignment function ranks candidate objects by maximizing similarity to positive cues while penalizing rejected ones, enabling fine-grained disambiguation in cluttered scenes. Our interactive framework provides substantial improvements in retrieval accuracy without additional supervision. On LVIS, IntRec achieves 35.4 AP, outperforming OVMR, CoDet, and CAKE by +2.3, +3.7, and +0.5, respectively. On the challenging LVIS-Ambiguous benchmark, it improves performance by +7.9 AP over its one-shot baseline after a single corrective feedback, with less than 30 ms of added latency per interaction.
\end{abstract}


\newcommand{\BibTeX}{\rm B\kern-.05em{\sc i\kern-.025em b}\kern-.08em\TeX}

\begin{document}


\pagestyle{fancy}
\fancyhead{}


\maketitle 


\section{Introduction}
Modern visual systems are increasingly expected to understand complex user intent to retrieve specific objects within open-world environments \cite{liu2024grounding, fu2025llmdet, weller2025theoretical}. This capability, termed object retrieval, requires the precise localization of a target based on varying linguistic or visual inputs \cite{kaul2023multi, ma2024ovmr, nara2024revisiting}. Unlike traditional object detection, which predicts  instances from a fixed category set  \cite{liu2024grounding}, object retrieval focuses on the specific instance that aligns with a user's intent. This distinction is foundational for interactive applications such as human-robot collaboration, AR/VR assistance, and advanced visual search.
Recent progress in open-vocabulary detection \cite{gu2021open, huang2024open, zareapoor2024learning} and visual grounding \cite{li2022grounded} have enabled models to transcend fixed label sets by aligning image and text embeddings in a shared semantic space.  However, these models operate under a one-shot retrieval design: a single query is matched to candidate regions, and the top-scoring region is returned as the prediction. This design is inherently limited. When queries are ambiguous or fine-grained (e.g., "the smaller umbrella with a floral pattern"), one-shot approaches produce incorrect or inconsistent predictions \cite{yao2023detclipv2, ma2024ovmr, huang2024open}.
Recent efforts like OVMR \cite{ma2024ovmr} and MMOVD \cite{kaul2023multi}, attempt to mitigate this ambiguity by incorporating multimodal cues, yet they still struggle in cluttered scenes where user prompts are vague or underspecified. 
Formally, given a single text query $T$ and a set of candidate image regions $R = \{r_1, \ldots, r_M\}$, standard open-vocabulary detectors  identify the most likely target region by computing the similarity between the query embedding \(E_T(T)\) and each region feature \(r_j\). The region with the highest similarity score is selected as the predicted object, $ b^\star = f_{\text{standard}}(T, R) = \text{argmax}\left( \text{sim}(E_T(T), r_j) \right)$. 

\begin{figure}[t]
        \centering
        \includegraphics[width=0.91\linewidth]{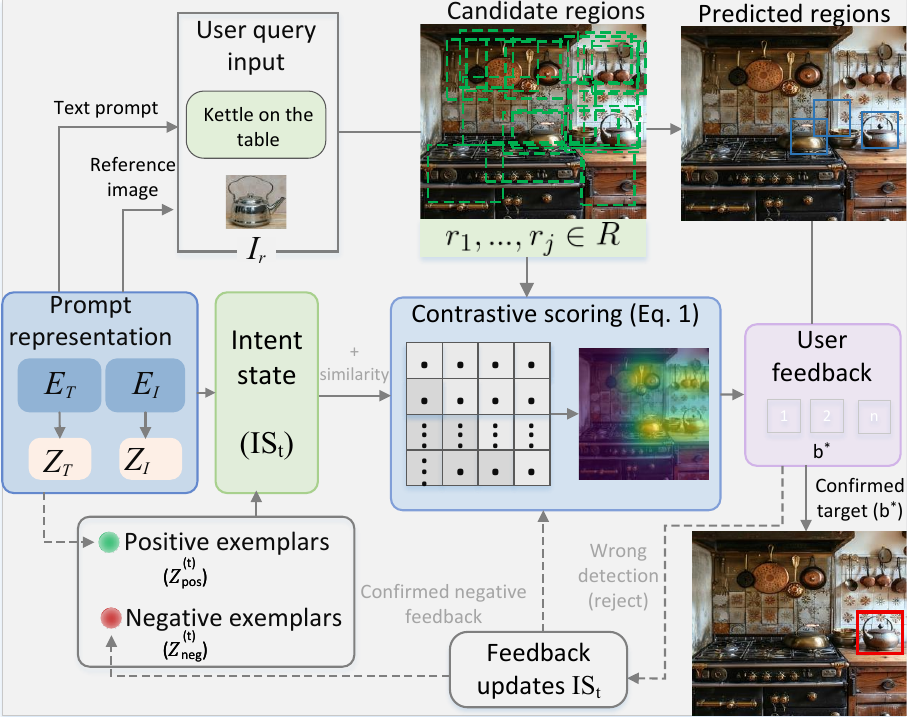}
        \caption{Overview of the proposed IntRec. It identifies a user-specified object through an interactive loop. An initial query is encoded to initialize the intent state, which containes both positive exemplars  \((Z_{pos}^{(t)})\) and negative exemplars $(Z_{neg}^{(t)})$. This state guides the contrastive scoring module, which ranks all candidate regions in the target image based on their similarity to the positive exemplars and dissimilarity to the negative ones. The feedback updates the intent state, enabling the model to accurately localize the target object \((b^\star)\). }
        \label{arch}
    \end{figure}

This retrieval function is stateless, its output depends only on the query embedding and region features,  with no mechanism to incorporate user feedback. In practice, these systems act as one-shot matchers that lack the temporal or logic-based depth to resolve ambiguity, a shortcoming that mirrors distribution misalignment issues found in generative diffusion tasks \cite{shamsolmoali2025missing}. This failure is most evident in cluttered scenes containing distractors (i.e., competing objects that are visually similar to the target). For example, a prompt like "the smaller red car" may result in multiple visually similar objects, making it difficult for the model to determine which object the user actually intended. %
%
%
%
To address this limitation, we propose Intent-based Retrieval (IntRec), an interactive and stateful framework for open-world object localization. At the core of IntRec is the Intent State (IS), a memory structure that accumulates both positive anchors (user-confirmed cues) and negative constraints (explicit rejections). A contrastive ranking function then evaluates candidate regions by maximizing similarity to positive anchors while penalizing similarity to negative ones. This mechanism enables fine-grained disambiguation, allowing the model to distinguish between highly similar objects after even a single corrective feedback. 
As illustrated in Figure~\ref{arch}, IntRec operates through a multi-turn interaction loop, where user feedback updates the intent state. Through this process, the model progressively aligns with the user intent, resolving fine-grained ambiguities that one-shot detectors cannot handle. We evaluate IntRec on large-scale open-vocabulary detection benchmarks, comparing against state-of-the-art baselines using AP for overall performance and Rare-AP for novel object detection. 
\begin{itemize}
   \item We formulate object retrieval as an interactive intent refinement problem, addressing the ambiguity limitations of open-vocabulary detectors.
   
\item We propose an Intent State module that accumulates both positive anchors and negative constraints from user feedback. A contrastive ranking function uses this state to refine retrieval and disambiguate fine-grained targets.

\item Extensive experiments show our model consistently outperforms state-of-the-art methods.
\end{itemize}







\section{Related Work}

\subsection{Open-Vocabulary Detection and Grounding}
Traditional object detectors rely on a fixed set of predefined labels, which limits their ability to recognize unseen categories at test time. To overcome this, open-vocabulary detection (OVDet) emerged, aiming to localize objects described by arbitrary textual queries. Early foundational works, such as ViLD \cite{gu2021open} and OWL-ViT \cite{minderer2022simple} demonstrated that large-scale vision-language models like CLIP \cite{radford2021learning} could be distilled into detection frameworks to enable zero-shot generalization. Subsequent research has focused on enhancing this transferability through improved distillation strategies and domain-adaptive designs \cite{wang2024ov, zareapoor2024learning, liu2025doga}. Despite their success in recognizing novel classes, these models are stateless, producing independent predictions per query without the capacity to resolve linguistic ambiguity or incorporate iterative user feedback. 
Parallel to OVDet, visual grounding focuses on the more granular task of localizing specific image regions corresponding to complex natural language phrases. GLIP \cite{li2022grounded} pioneered the unification of detection and phrase grounding by reformulating object detection as a word-region alignment task. 
This direction was further advanced by transformer-based architectures like Grounding DINO \cite{liu2024grounding}, DetCLIPv2 \cite{yao2023detclipv2}, and CoDet \cite{ma2024codet}. 
Recent models continue to refine these approaches using query mechanisms \cite{fu2025llmdet, jin2024llms, zareapoor2025bimac}, semantic graph constraints \cite{lin2025sgc}, and category-specific knowledge distillation \cite{ma2025cake}, allowing detectors to better understand nuanced textual descriptions and object-to-object relationships. MMOVD \cite{kaul2023multi} and OVMR \cite{ma2024ovmr} extended these ideas with multimodal fusion, improving text-image interactions. 
While these models excel at category-level recognition, they struggle with one-of-many scenarios where multiple candidate objects share near-identical semantic features. In these cases, models often assign nearly identical confidence scores to all matching instances, failing to detect the specific object intended by the user. 
This limitation is particularly evident in examples where a model must identify one among several similar objects (e.g., second-third columns of Figure~\ref{codet}). Such ambiguity highlights a shortcoming of current open-vocabulary and grounding approaches.


\begin{table}\small
\centering
\begin{tabular}{l|p{0.67\linewidth}} 
\toprule
\textbf{Symbol} & \textbf{Definition} \\ 
\midrule
\(I_s, I_r\)        & Target image used for retrieval, and an optional user-provided reference image. \\ 
\(p_0\)             & Initial multi-modal prompt, consisting of a text description \(T_0\) and/or a reference image \(I_r\). \\
\(b, b^\star\)      & Candidate bounding box and the final confirmed target object. \\
\(R = \{r_1, \ldots, r_M\}\) & Set of \(M\) candidate region feature vectors extracted from \(I_s\). \\
\(E_T, E_I\)        & Pre-trained text and image encoders (e.g., CLIP). \\ 
\(z_T, z_I, z_p\)   & Embeddings of the text prompt, reference image, and fused query representation. \\
\(IS_t\)            & Intent State at interaction turn \(t\). \\
\(Z_{pos}^{(t)}\)   & Positive exemplar within \(IS_t\); stores embeddings representing user-confirmed cues. \\
\(Z_{neg}^{(t)}\)   & Negative exemplar within \(IS_t\); stores embeddings representing rejected. \\
\(f_t = (b_j, s_t)\) & Feedback tuple, where \(b_j\) is the selected region and \(s_t\!\in\!\{\text{positive}, \text{negative}\}\) is the feedback type. \\
\bottomrule
\end{tabular}
\caption{Notation summary for the proposed IntRec.}
\label{notation}
\end{table}

\subsection{Interactive Modeling}
The idea of learning from user interaction has long been explored in content-based retrieval \cite{rui1999image, harman1992relevance, nara2024revisiting}. In these systems, users provide relevance feedback by marking retrieved results as positive or negative, allowing the model to iteratively refine the query representation, often using the Rocchio algorithm \cite{rocchio1971relevance}. These early methods demonstrated the power of feedback for refining retrieval results but were not designed for object localization within a single image. More recent work has introduced interaction into visual models in different ways. Diffusion-TTA \cite{prabhudesai2023diffusion} adapts vision encoders using test-time feedback, while interactive segmentation methods \cite{sofiiuk2022reviving} show how user corrections can refine object boundaries. Closest to our work, Qin et al. \cite{qin2025human} proposed an interactive learning framework for text-to-image person re-identification. However, while effective for identity-matching across image galleries, such methods are not optimized for resolving the semantic ambiguity of general object categories in cluttered, open-world scenes. A common limitation persists across these domains: predictions are generally made independently for each query, with no mechanism to resolve distractors through a stateful dialogue. Our work addresses this gap by proposing IntRec that combines vision-language models with a contrastive feedback mechanism for user-guided object localization. 


\section{Proposed Model}
Given a target image \(I_s\), our goal is to localize a specific object \(b^\star\) through a sequence of user interactions \(t=0,1, \ldots, K\), rather than relying on predefined categories. At the initial turn \(t=0\), the user provides a query prompt \(p_0\), which may include a text description $T_0$, a reference image $I_r$, or both. The model extracts a set of $M$ candidate regions from \(I_s\), represented by their feature embeddings \(R=\{r_1,\ldots, r_M\}\), where each \(r_i\in R^d\). At each subsequent turn \(t\!>\!0\), the user provides feedback \(f_t=(b_j,s_t)\), where \(b_j\) is a candidate region and \(s_t\in\{\text{positive}, \text{negative}\}\) indicates  whether that region matches the intended target. Positive feedback confirms a user's target object, while negative specifies visual attributes to avoid. Ambiguity arises when multiple regions satisfy the initial query equally well, making user feedback essential for disambiguation.


\subsection{Intent State (IS) Representation}
In most retrieval models, user intent is represented as a single embedding vector obtained by encoding the input query. While simple, this formulation compresses all semantic cues into a single point and lacks a mechanism to learn from user rejections \cite{weller2025theoretical}, limiting its ability to distinguish between similar candidates.  
We propose an Intent State (IS), a memory representation that evolves throughout the interaction. At any turn \(t\), the IS is defined as a tuple of two exemplar sets: $IS\textsubscript{t}={Z}_{pos}^{(t)}, {Z}_{neg}^{(t)}$, where the first set stores positive anchors (embeddings representing user-confirmed cues or desired attributes), and the second stores negative constraints (embeddings of rejected attributes). The process begins by converting the initial prompt \(p_0\) into the state \(IS_0\). The text prompt \(T_0\) and reference image $I_r$ are encoded using CLIP encoders as \(z_{T}=E_T(T_0)\) and \(z_{I}=E_I (I_r)\), respectively. These are fused into a single query vector $z_p^{(0)}=\alpha\cdot z_T+(1-\alpha)\cdot z_I$, where $\alpha$ is a balancing hyperparameter.  This vector forms the first entry in the positive exemplar set, initializing the state as  ${Z}_{pos}^{(0)}=\{z_p^{(0)}\}$, and ${Z}_{neg}^{(0)}=\emptyset$. By modeling intent as a distribution of exemplars rather than a single embedding, this representation enables more precise and stable retrieval.


\subsection{Candidate Ranking Function}
At each turn \(t\), the model ranks all candidate regions \(R\) according to the current state IS\textsubscript{t}. To achieve this, we propose a contrastive alignment score that acts as a exemplar-based classifier, measuring each candidate's relationship to both positive and negative exemplars. Formally, the score for a candidate region \(r_j\) is defined as 
\begin{equation}
(r_j\mid IS_t)= \underbrace{\max_{z^+ \in {Z}_{pos}^{(t)}} \text{cos}(r_j, z^+)}_{\text{Similarity to positive intent}} - \underbrace{\lambda \cdot \max_{z^-\in {Z}_{neg}^{(t)}} \text{cos}(r_j, z^-)}_{\text{Penalty for negative intent}}
\label{eq:score}
\end{equation}
where $\lambda$ controls the influence of negative constraints. The first term promotes regions that align closely with any positive exemplar. For instance, if $z^+$  contains embeddings for both a text prompt and a reference image, the model identifies candidate regions that match either cue. The second term penalizes regions similar to rejected exemplars \((z^{-})\), effectively creating low-scoring valleys around non-target concepts in the embedding space. Together, these terms enable fine-grained discrimination between visually similar objects.

\subsection{Interactive State Update}
The interactive state update is the core learning mechanism of our framework, allowing the model to learn from corrections and progressively converge on the intended target. Let \(r_j\) be  the embedding of the region selected by the user at turn $t+1$ with feedback \(f_{t+1}=(b_j, s_{t+1})\). The update rule depends on the feedback type. (i) Negative feedback ($s_{t+1}=\text{negative}$): the rejected region's feature vector is added to the negative set $(r_j \to {Z}_{neg})$
\begin{equation}
  {Z}_{neg}^{(t+1)} = {Z}_{neg}^{(t)} \cup \{r_{j}\}  
\end{equation}
while the positive set remains unchanged  ${Z}_{pos}^{(t+1)}={Z}_{pos}^{(t)}$. This update teaches the model which visual features to avoid (e.g., distractors).  (ii) Positive feedback $(s_{t+1}=\text{positive})$: if the user confirms a region (object) or provides a new textual prompt (encoded as \(z_{\text{new}}\)), the corresponding feature is added to the positive exemplar 
\begin{equation}\begin{aligned}
{Z}_{pos}^{(t+1)}&= {Z}_{pos}^{(t)} \cup \{r_j, z_{new}\} \\
{Z}_{neg}^{(t+1)}&={Z}_{neg}^{(t)} \quad \text{(Unchanged)} 
 \end{aligned}
\end{equation}
the feedback feature can be an image region \(r_j\), or the encoded new text prompt. The updated $IS_{t+1}=({Z}_{pos}^{(t+1)}, {Z}_{neg}^{(t+1)})$ is then used in the next turn to re-score the candidates, using Eq. \ref{eq:score}. This loop continues until the target is confirmed. Through this process, the model learns not only which object is wrong, but also which visual features are misleading, an ability absent from most open-world detectors. 
In summary, IntRec begins with a query encoded into text and image embeddings, forming the initial intent state (IS\textsubscript{0}). The model generates candidate regions for the target image and evaluates them using the contrastive scoring function (Eq.~\ref{eq:score}). The top-ranked region is displayed to the user, who provides feedback to confirm or reject the result. This feedback updates the intent state (IS\textsubscript{t}$\rightarrow$IS\textsubscript{t+1}), progressively refining subsequent predictions. The full procedure is summarized in Algorithm \ref{algo}, Table \ref{notation}.

\subsection{Theoretical analysis of ambiguity resolution} \label{analyse}
Most open-vocabulary detectors (e.g., OWL-ViT, Grounding DINO), simply  returns the region (object of interest, $r^\star$) with the highest similarity to the query $T$, and defined by
\begin{equation}
b^\star = f_{\text{standard}}(T, R) = \underset{b_j}{\text{argmax}} (\text{sim}(E_T(T), r_j))
\end{equation}
An ambiguity condition exists if there is at least one distractor object $r_d\neq r^\star$, such that its similarity to the query is greater than or equal to the true target's similarity $\text{sim}(E_T(T), r_d) \ge \text{sim}(E_T(T), r^\star)$. Under this condition, the stateless model will incorrectly select $r_d$ and has no mechanism for correction. 
In contrast, our framework is designed to resolve such ambiguity.  At the initial turn $(t=0)$, our score function behaves identically to the standard case \(S(r_j \mid IS_0) = \text{sim}(r_j, E_T(T))\), and may also select the distractor $r_d$. After receiving negative feedback on $r_d$, the intent state updates to $IS_1 = (Z_{pos}^{(1)}, Z_{neg}^{(1)}) = (\{E_T(T)\}, \{r_d\})$. The new scores for the distractor and the target become
\(
S(r_d|IS_1) = \text{sim}(r_d, E_T(T)) - \lambda, \quad
S(r^\star|IS_1) = \text{sim}(r^\star, E_T(T)) - \lambda \text{sim}(r^\star, r_d)
\)
The ambiguity is resolved if $S(r^\star | IS_1) > S(r_d | IS_1)$, which holds
\[
\lambda (1 - \text{sim}(r^\star, r_d)) >
\text{sim}(r_d, E_T(T)) - \text{sim}(r^\star, E_T(T))
\]
As the right-hand side is a small non-negative, and $1-\text{sim}(r^\star, r_d)$ is a positive constant (since $r^\star\neq r_d$), the inequality can always be satisfied with a suitable penalty weight $\lambda$. This proves that our contrastive mechanism is guaranteed to resolve the ambiguity. Consider a query for "the most front cup" in a scene with multiple similar cups. A stateless model may correctly identify a cup in the front row but select the wrong one, as several candidates have nearly identical high scores. However, our interactive model, after detecting the wrong object, follow-up by "not this, the cup near the left bottle". The model then, parses "not this" as a negative constraint. Using this feedback, we update both sets of the intent state simultaneously. The visual features of the rejected cup are added to the negative constraint set $(Z_{neg})$, while the new clue, is encoded and added to the positive anchor set $(Z_{pos})$. In the subsequent search, the model re-ranks all candidates based on this richer, more constrained intent state. This corrective feedback and update both positive and negative constraints is what separates our model from current open-vocabulary detectors. 


\begin{algorithm}[t]\small
\caption{Our proposed dynamic intent search algorithm}
\label{algo}
\begin{algorithmic}
\Require Target image \(I_s\), prompt $p_0=\{T_0, I_r\}$, max interaction turns $K$.
\Ensure Localized target \(b^\star\).
\State Extract candidate region embeddings $R = \{r_1, \dots, r_M\}$ from $I_s$
\State Encode query $z_T = E_T(T_0)$, $z_I = E_I(I_r)$, $z_p^{(0)} = \alpha z_T + (1-\alpha) z_I$
\State Initialize intent state: ${Z}_{pos}^{(0)} = \{z_p^{(0)}\}$, ${Z}_{neg}^{(0)} = \emptyset$

\For{$t = 0, 1, \dots, K-1$}
    \For{$r_j \in R$}
        \State $S_{pos} = \max_{z^+ \in {Z}_{pos}^{(t)}} \cos(r_j, z^+)$
        \State $S_{neg} = \max_{z^- \in {Z}_{neg}^{(t)}} \cos(r_j, z^-)$ \textbf{if} ${Z}_{neg}^{(t)} \neq \emptyset$
        \State $S(r_j) = S_{pos} - \lambda S_{neg}$ \Comment{Contrastive ranking (Eq.~\ref{eq:score})}
    \EndFor
    \State Present top-$k$ candidates to the user and receive feedback $f_{t+1} = (b_j, s_{t+1})$
    \If{$s_{t+1} = \text{positive-confirmation}$}
        \State \Return $b^\star = b_j$
    \ElsIf{$s_{t+1} = \text{positive-refinement}$}
        \State ${Z}_{pos}^{(t+1)} = {Z}_{pos}^{(t)} \cup \{r_j \text{ or } z_{\text{new}}\}$
    \ElsIf{$s_{t+1} = \text{negative}$}
        \State ${Z}_{neg}^{(t+1)} = {Z}_{neg}^{(t)} \cup \{r_j\}$
    \EndIf
\EndFor
\end{algorithmic}
\end{algorithm}

\section{Experiments}\label{exp}
\subsection{Architecture and Implementation Details}
We use the pre-trained, frozen CLIP ViT-B/16 encoders \cite{radford2021learning} for both text \((E_T)\) and image \((E_I)\) feature extraction. All textual prompts and image region features are projected into a L2-normalized embedding space of dimension d=512. To generate object proposals (candidate regions), we follow recent works \cite{zhou2022detecting, ma2024ovmr, kaul2023multi} and employ CenterNet2 \cite{zhou2021probabilistic} with a ResNet-50 backbone pre-trained on the ImageNet-21k dataset \cite{ridnik2021imagenet}. The weights of both the CLIP and CenterNet2 models are kept frozen during inference. For any target image \(I_s\), we use the detector's class-agnostic proposal to generate \(M=100\) candidate bounding boxes. The features for these proposals, \(R=\{r_1,\ldots,r_{100}\}\), are then extracted using frozen CLIP image encoder. To measure how the interaction process improves detection performance, we adopt the following evaluation protocol. The interaction is simulated for a maximum of \(K=2\) turns. Turn-0 (baseline): for each query, we first evaluate the model's initial, non-interactive prediction, and compute the AP @ Turn-0 score. Turn-1 (post-feedback): if the model's top-1 prediction at Turn-0 was incorrect, we provide negative feedback on that specific object. The model updates its intent state and generates a refined ranked list of objects. We then compute the AP @ Turn-1. 
The hyperparameters are set as: $\alpha \ \text{(modality fusion)}=0.6$, \ $\lambda \ \text{(negative penalty weight)} = 1.0$, \ and $M$ (number of candidate regions) $=100$. 


\subsection{Dataset and Evaluation metrics}\label{benchmark}
We evaluate our framework on two large-scale object detection datasets: LVIS v1 \cite{gupta2019lvis} and Objects365 \cite{shao2019objects365}. Following the ViLD protocol \cite{gu2021open}, we treat the 866 common and frequent LVIS categories as base classes, and the 337 rare categories as novel classes to assess open-vocabulary generalization. Objects365, which contains 365 categories across more than 600k images, is used for transfer detection experiments. We report mean Average Precision (AP) across all categories, as well as AP(r), AP(c), and AP(f) for rare, common, and frequent subsets, respectively. Our goal is to improve AP(r), without significantly reducing overall AP. 
To specifically evaluate our model's ability to handle ambiguous retrieval scenarios, we constructed a challenging subset from the LVIS, which we call LVIS-Ambiguous. This benchmark is designed to include cases where multiple visually similar objects of the same category appear in the same image, causing standard open-vocabulary detectors to make incorrect predictions.  We used a pre-trained Grounding DINO \cite{liu2024grounding} as a probe. For each ground-truth object (in an image $I$) with bounding box $b_{gt}$ and category label $c_{gt}$, we generate a simple category-level text prompt (i.e,  $T$=a photo of a $c_{gt}$). This triplet $(I, T, b_{gt})$ forms one candidate sample for the benchmark. We then input 
\(I\) and \(T\) into the frozen Grounding DINO, which outputs a ranked list of \(N\) detected objects with bounding boxes and confidence scores, $D=[(b_1, s_1), (b_2,s_2),\ldots,(b_N,s_N)]$. Next, we compute the IoU between the ground-truth box $b_{gt}$ and every predicted box $b_i\in D$, and identify $b_{gt-pred}$ with the highest IoU to $b_{gt}$ (let its rank in the list be $k$). We then inspect all predictions ranked higher than this true target (i.e., $b_j$ with rank $j<k$). A prediction $b_j$  is a distractor  if it satisfies both conditions: it has a low overlap with the true object (i.e., \(IoU(b_j, b_{gt}) < 0.5\)), and, it  belongs to the same category $c_{gt}$ as the true object (verified via ground-truth annotations). If at least one such distractor is ranked higher than the true target, the sample  $(I, T, b_{gt})$ is an ambiguous case, and added to the LVIS-Ambiguous benchmark. This procedure yields a curated subset of fine-grained, visually confusing cases.

For this experiment we report two scores for our model. In AP @ Turn-0, we first run our model in a non-interactive mode, using only the initial prompt (without using feedback), then calculate the AP score.  In AP @ Turn-1, for each sample, we take the top-1 prediction from the Turn-0 and provide it as a negative feedback signal. Our model updates its intent state and produces a new, refined ranked list. We calculate the AP score on this new list.

\begin{table}
\centering
\begin{tabular}{l|c|l}
\toprule
Method     &  backbone     &    AP (r / c / f)    \\
\midrule
Detic \cite{zhou2022detecting}  & ResNet-50  &   27.4 (18.6 \color{gray!50}{/ - / -})   \\
DetPro \cite{du2022learning} & ResNet-50 & 26.8 (20.3 \color{gray!50}{/ 26.5 / 28.9})   \\
RegionCLIP \cite{zhong2022regionclip}  &  ResNet-50   & 28.2 (17.1 \color{gray!50}{/ 27.4 / 34.0})  \\
ViLD \cite{gu2021open} & ResNet-50  &  25.8 (16.6 \color{gray!50}{/ 24.6 / 30.1}) \\
CCKT-Det \cite{zhang2025cyclic} & ResNet-50 & 27.1 (18.2 \color{gray!50}{/ - / -})  \\   
F-VLM \cite{kuo2022f} & ResNet-50&   24.2 (18.6 \color{gray!50}{/ - / -})    \\
BARON \cite{wu2023aligning} & ResNet-50 &    29.5 (23.2 \color{gray!50}{/ 29.6 / 33.8})   \\
VLDet \cite{lin2023learning} & ResNet-50 & 30.1 (21.7 \color{gray!50}{/ 29.8 / 34.3})   \\
DVDet \cite{jin2024llms} & ResNet-50 &   31.2 (23.1 \color{gray!50}{/ 31.2 / 35.4})   \\
NRAA \cite{qiang2024open} & ResNet-50 & 27.8 (21.2 \color{gray!50}{/ 27.0 / 31.5}) \\  
MMOVD \cite{kaul2023multi} & ResNet-50 &   31.3 (19.8 \color{gray!50}{/ 31.5 / 34.2})   \\
CoDet \cite{ma2024codet} & ResNet-50 & 31.7 (24.5 \color{gray!50}{/ 31.0 / 35.4})    \\
LBP \cite{li2024learning} & ResNet-50 & 29.9 (24.7 \color{gray!50}{/ 29.5 / 32.8}) \\
OVMR \cite{ma2024ovmr} &  ResNet-50 &   33.1 (23.5 \color{gray!50}{/ 32.1 / 36.9})   \\
CAKE \cite{ma2025cake} & ResNet-50 &  34.9 (25.0 \color{gray!50}{/ 34.8 / 38.4})    \\
MIC \cite{wang2024open} & ResNet-50 & 33.8 (22.1 \color{gray!50}{/ 33.9 / 40.0}) \\  
\midrule
\rowcolor{lightgray!30}
IntRec- T &  ResNet-50   &  \bf 35.0 ({{24.7}} \color{gray!50}{/ 34.1 / 38.0})  \\ 
\rowcolor{lightgray!30}
IntRec- MM  &  ResNet-50  &   \bf{35.4} ({{25.6}} \color{gray!50}{/ 34.7 / 38.2 })  \\ 
\bottomrule
\end{tabular}
\caption{Comparison results on LVIS using ResNet-50 detector backbone. AP(r) and AP are the main evaluation metric for this experiment. Rows for our models are highlighted, representing results from text, vision, and multimodal.  }
\label{tab1}
\end{table}


\subsection{Comparison with State-of-the-Art Methods} \label{sec4}
\subsubsection{Open world object detection}
We evaluate our model on the LVIS and compare it against recent state-of-the-art open-vocabulary detectors. To ensure a fair comparison, all models use a ResNet-50 \cite{he2016deep} backbone. Performance is reported in Table \ref{tab1} using mean Average Precision (AP), with a focus on rare classes AP\textsubscript{r}. While methods such as MMOVD \cite{kaul2023multi}, MIC \cite{wang2024open}, and OVMR \cite{ma2024ovmr} mitigate ambiguity by using multiple (e.g., 10) text and image prompts, our model takes a simpler strategy. Given a single query, our model performs an initial detection pass to identify the top-1 ranked object, which serves as a distractor. It then performs a second inference pass, where all candidate regions are re-ranked using our contrastive scoring function, with the original query as a positive anchor and the top-1 prediction from the first pass as a negative constraint. The final ranked list from this second pass is used to compute our model's AP scores.
As shown in Table \ref{tab1}, our text-only model sets a new state-of-the-art on LVIS. When augmented with a reference image, performance improves further, surpassing the previous best methods on AP\textsubscript{r}, including CAKE (25.0), CoDet (24.5), LBP (24.7). Although some recent models, such as MM-GDINO \cite{zhao2024open}, Grounding DINO \cite{liu2024grounding}, LLMDet \cite{fu2025llmdet}, DesCo \cite{li2023desco}, and DITO \cite{kim2024region}, using larger backbones (e.g., Swin-T and ViT-S/16), their AP(r) remains limited (23.6 / 10.1 / 26.0 / 19.6 / 26.2, respectively), and they require significantly more computational resources.

\begin{table}
  \centering
  \begin{tabular}{l|ccc|ccc}
    \toprule
  Method &  \multicolumn{3}{c|}{object365} & \multicolumn{3}{c}{COCO}  \\ \cmidrule{2-7}
  &    AP & AP-50 & AP(r) & AP & AP-50 & AP-75  \\
       \midrule
DetPro \cite{du2022learning} & 11.7 & 18.2 & 9.8 & 34.9 & 53.8 & 37.4
\\

BARON \cite{wu2023aligning} & 12.7 & 20.3 & 10.2 & 36.2 & 55.7 & 39.1   \\

CoDet \cite{ma2024codet}& 13.8 & 20.6 & 10.5 & 37.4 & 55.1 & 39.5   \\

CCKT-Det \cite{zhang2025cyclic} & 13.4 & 19.7 & 11.5 & 38.5 & 53.2 & 42.1 \\

OVMR \cite{ma2024ovmr}  & 12.3 & 19.5 & 10.7 & 37.6 & 54.7  & 40.3 \\

DetLH \cite{huang2024open} & 14.6 & 21.2 & 11.9 & 38.9 & 56.5 & 41.6   \\
\midrule
\rowcolor{lightgray!30}
IntRec (Turn-0) &  13.8 & 21.4 & 11.5 & 38.6 & 56.0 & 41.3 \\ \rowcolor{lightgray!30}
IntRec (Turn-1) & \bf17.2 & \bf 23.9 & \bf14.7 & \bf41.5 & \bf57.8 & \bf43.7     \\
  \bottomrule
  \end{tabular}
  \caption{Transfer detection result. All models are trained on the LVIS/ImageNet-21K, and evaluated on Object365 and COCO. Our model is shown at both Turn-0 and Turn-1. }
  \label{tab2}  
  \vspace{-10pt}
\end{table}

\begin{table}
 \centering
  \begin{tabular}{c|c} \toprule
 Model & AP on LVIS-Ambiguous \\  \midrule
CoDet \cite{ma2024codet} & 13.9    \\
OVMR \cite{ma2024ovmr}	& 14.5	   \\
\midrule
\rowcolor{lightgray!30}
IntRec \ (Turn-0)	        & 14.8	 \\
\rowcolor{lightgray!30}
IntRec \ (Turn-1)	         & \bf22.7 \\
\bottomrule
    \end{tabular}
    \caption{Performance on the LVIS-Ambiguous Benchmark. While existing state-of-the-art models perform well on LVIS overall, their performance collapses under ambiguity. In this challenging scenario, our interactive model demonstrates a substantial recovery and clear performance advantage.}
    \label{ambigous}
\end{table}


\begin{figure*} 
    \centering
  \includegraphics[width=0.87\textwidth]{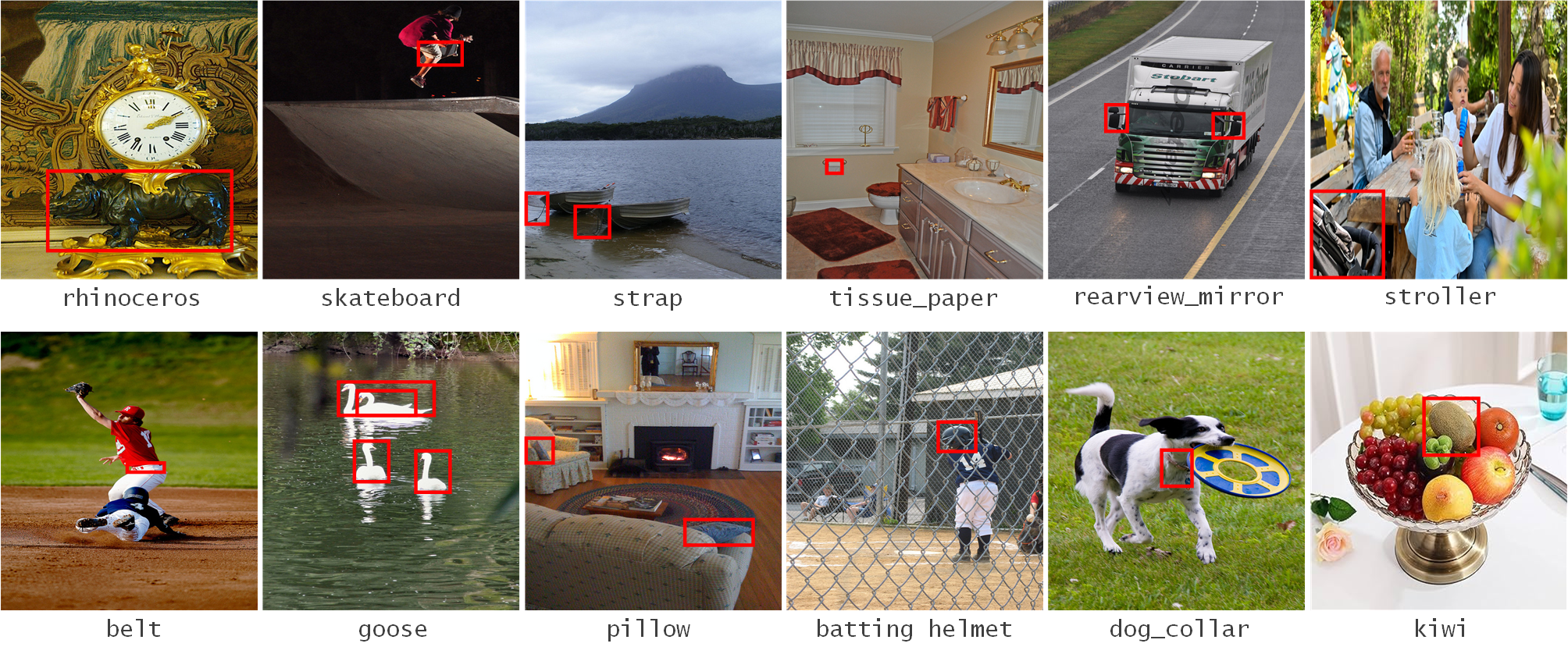}  
  \   \caption{Qualitative examples of our proposed model detecting rare categories in the LVIS validation set using textual prompt.}
   \label{fig3}
\end{figure*}

\subsubsection{Analysis on the LVIS-Ambiguous benchmark}
Our model is designed to recover from initial mispredictions through interaction. Consider an image densely packed with several small ceramic cups, where the target is "the cup at the very front". Most detectors, which relies on a single similarity score, often fail in this cases \cite{xi2025ow, wang2024ov, lin2025sgc}. Because all cups appear visually similar, their similarity scores are nearly identical, making the top-ranked prediction unstable and incorrect (see Section \ref{analyse}). With a simple feedback such as "not this one, the cup near the left bottle", our model updates its intent state to suppress the rejected region and refocus on the correct target. To evaluate this behavior, we use our LVIS-Ambiguous benchmark (see Section \ref{benchmark}). We compare the performance of top-performing baselines against our model in two modes: Turn-0, representing the initial non-interactive prediction, and Turn-1, representing the prediction after one round of feedback. The results in Table \ref{ambigous}, reveal that all models perform poorly at Turn-0, confirming the difficulty of this benchmark. However, after a single corrective interaction, our model's performance at Turn-1 rises to 22.7, an improvement of +7.9, demonstrating its ability to recover from initial mispredictions and handle visual ambiguity.


\subsubsection{Transfer detection setting}
To evaluate our model generalization capabilities, we test it in a zero-shot transfer detection setting. The model, which has been trained on LVIS/ImageNet-21k, is evaluated directly on the Objects365 and COCO datasets without any fine-tuning. AP(r) is evaluated on least frequent 25\% categories in the Objects365 training set. We evaluate our model in two modes: Turn-0 performance, where the prediction is based on the initial text prompt., and Turn-1 which is the result after applying our correction (feedback) mechanism, where the top-1 prediction from Turn-0 is used as a negative constraint to re-rank all candidates. The results are shown in Table \ref{tab2}. Our model at Turn-0 performs comparably with other baselines, but, after a single correction pass, our Turn-1 performance shows a significant boost across all metrics on both datasets, specially on the rare categories.

\begin{figure}[h]
    \centering
  \includegraphics[width=0.4\textwidth]{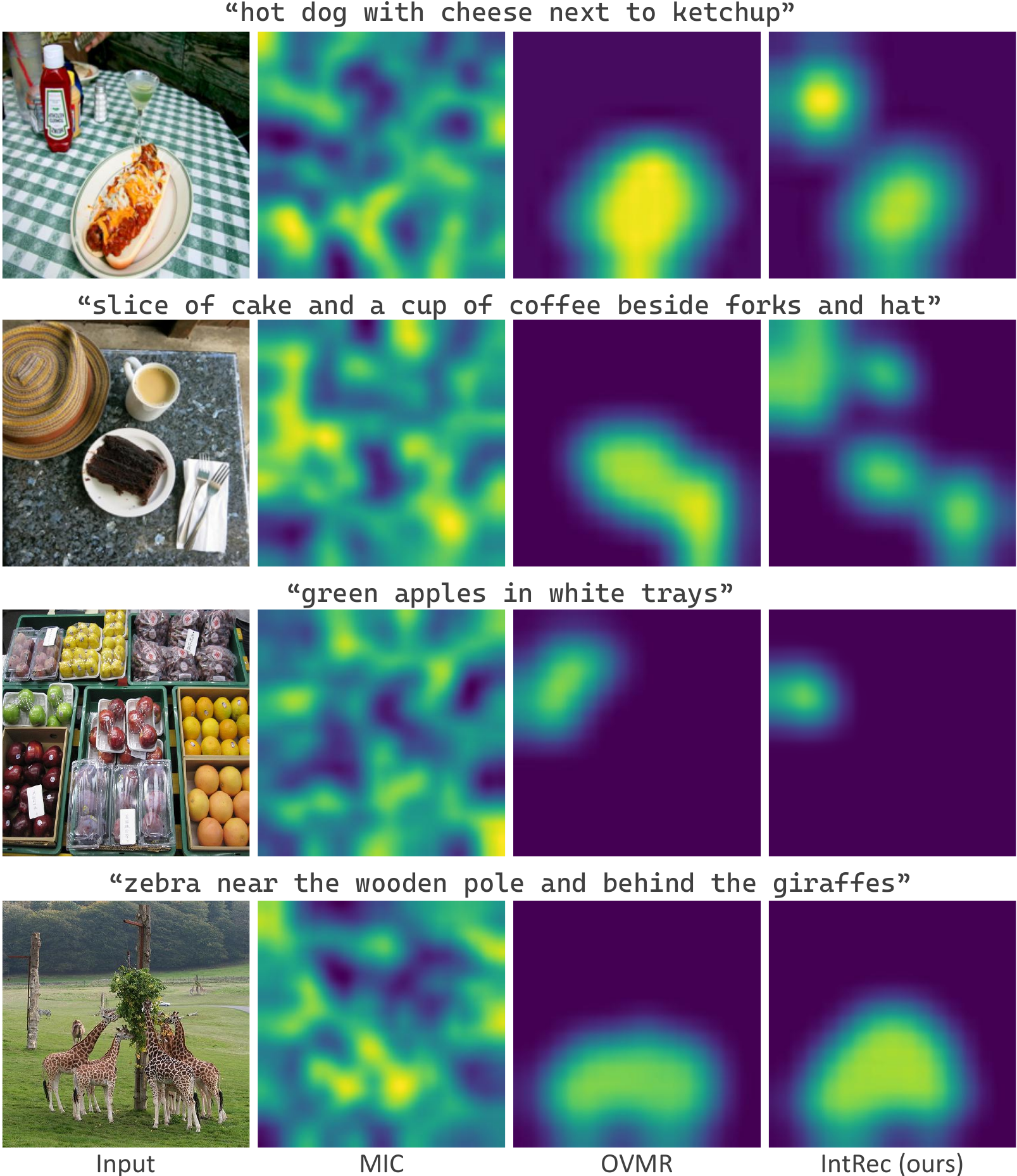}
    \caption{Localization comparison. While baseline models produce diffuse or imprecise heatmaps, our model generates sharp, accurate localizations that correctly ground all semantic components of the prompt. For instance, it correctly distinguishes the green apples from other fruit and localizes the zebra despite the presence of nearby giraffes.}
   \label{attention}
\end{figure}


\begin{figure*}
    \centering
    \includegraphics[width=0.88\linewidth]{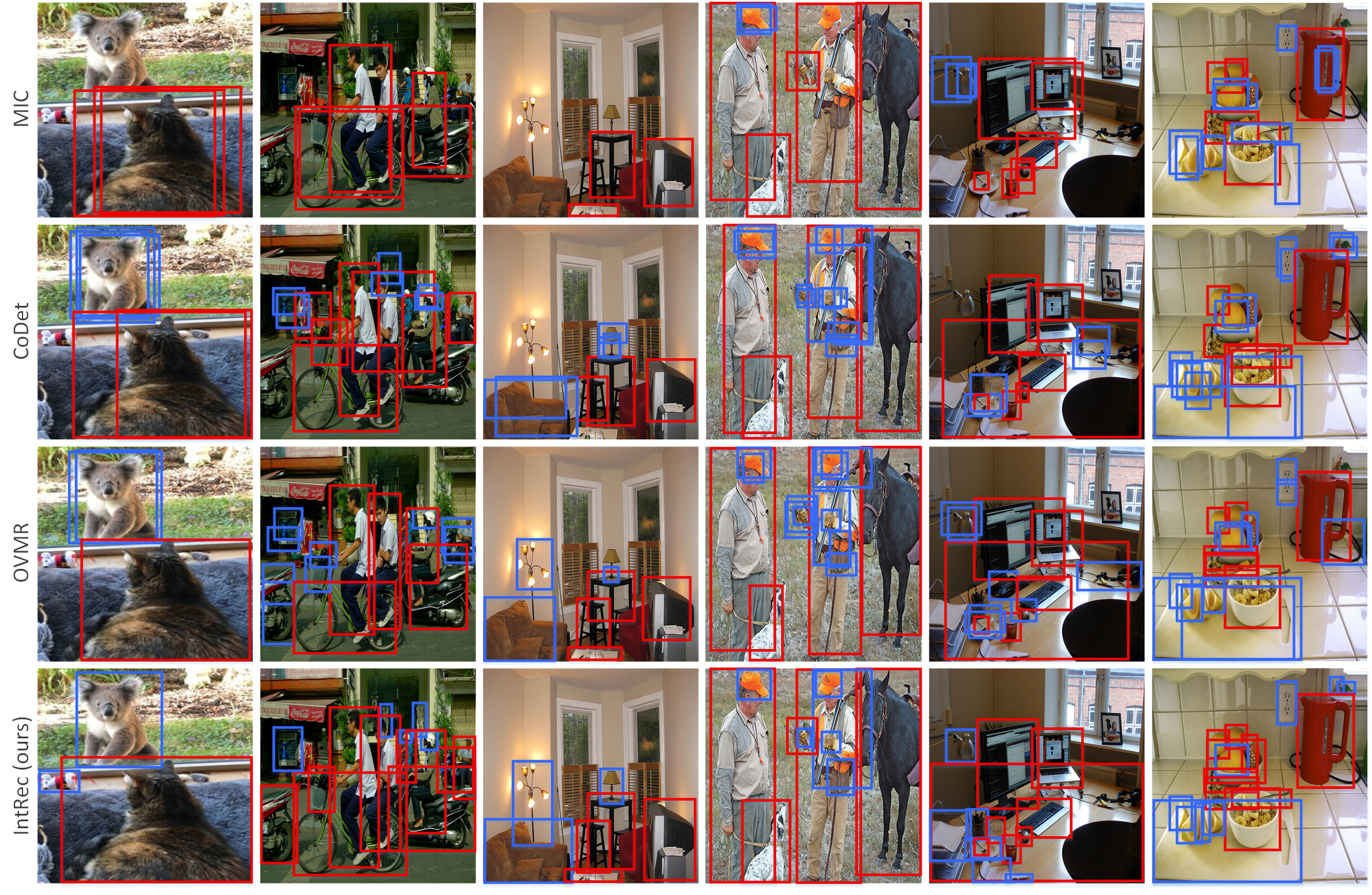}
     \caption{Detection comparison in cluttered scenes. Blue boxes indicate detections of novel/rare categories, red boxes indicate detections of known (base) categories. While the baseline models frequently produce a redundant and overlapping detections for novel objects, our model generates accurate predictions, suppress duplicate detections in dense environments. }
    \label{codet}
\end{figure*}

\subsubsection{Alignment strategy: Local vs. Global ranking}
\begin{figure}
    \centering
    \includegraphics[width=0.82\linewidth]{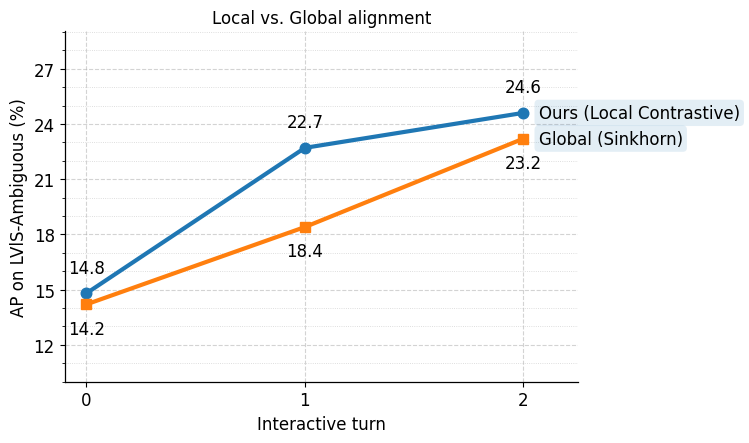}
    \caption{Analysis of Local vs. Global alignment on the LVIS-Ambiguous benchmark over two interactive turns. Both models start with a low AP at Turn 0 (initial prompt). However, after the first corrective feedback (Turn 1), our model increases by +7.9, outpacing the +4.3 gain of the global baseline.}
     \label{sinkhorn}
\end{figure}

One of the key mechanisms in our framework is the ranking function (Eq. \ref{eq:score}), which evaluates each object (candidate region $r_j$) independently using a local contrastive score. To justify whether this local strategy is indeed optimal, we compare it against a global assignment baseline based on optimal transport. For this baseline, we replace our scoring function at each interactive turn $t$ with Sinkhorn-based algorithm \cite{cuturi2013sinkhorn}, and construct a cost matrix between the positive anchors $Z_{pos}$ and all candidate regions $R$, where the cost is inversely proportional to their similarity. Negative constraints, $Z_{neg}$,  are used as a high cost (penalty) for undesirable regions in this matrix. The Sinkhorn algorithm then computes a soft assignment plan, from which the final ranking of candidates is derived. We evaluate both variants on the LVIS-Ambiguous benchmarks over a maximum of \(K=2\) interactive turns, focusing on novel/rare category performance. As shown in Figure \ref{sinkhorn}, our model consistently achieves a higher final AP, surpassing the Sinkhorn-based global assignment variant by +1.4 (at turn=2). We hypothesize that this advantage arises because local scoring can better exploit the sharp, localized signal from a negative constraint, whereas global assignment tends to dilute this signal across the entire candidate set.

\subsubsection{Qualitative results}
Figure \ref{attention} shows visual attention maps generated by our model for several complex queries, alongside results from MIC \cite{wang2024open} and OVMR \cite{ma2024ovmr}. These visualizations illustrate how well each model captures fine-grained linguistic attributes in the image. Our model produces sharper and more focused attention regions, indicating a stronger correspondence between the textual query and the visual evidence. Specially, the MIC model generates diffuse and scattered attention, indicating difficulty in grounding the full query. While OVMR can identify the correct object region, still displays coarse and imprecise focus. For example, in the query "green apples in white trays", both baselines fail to distinguish between visually similar objects, whereas our model provides a focused heatmap that aligns with the target objects described in the text (detection result for our model is given in Figure \ref{fig3}). 

To further illustrate this behavior, Figure \ref{codet} compares the final bounding box predictions of our model against several baselines. In these examples, red boxes indicate detections of base (known) categories, and blue boxes denote detections of novel/rare categories. Although the baselines are capable of detecting novel objects, they often produce redundant and overlapping boxes (i.e., multiple imprecise detections corresponding to the same instance). This is a failure mode for open-vocabulary detectors in dense scenes \cite{liu2024grounding, fu2025llmdet}.  In contrast, our model yields more accurate detections, producing well-localized boxes that correspond to the intended objects.

\begin{figure}
    \centering
    \includegraphics[width=0.92\linewidth]{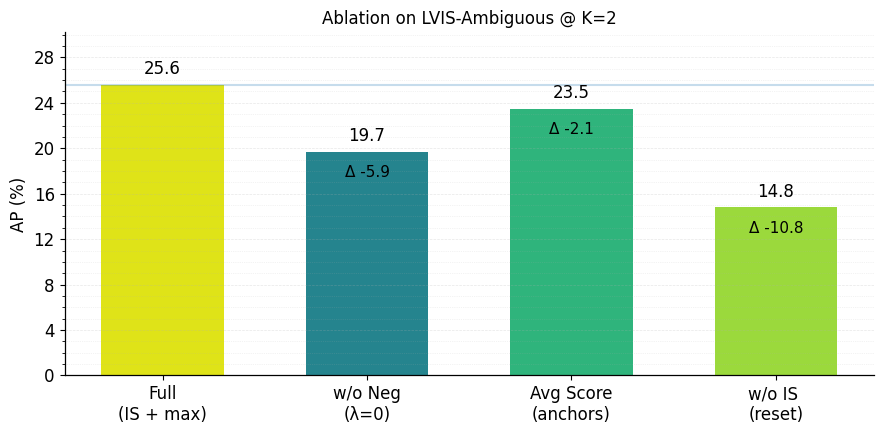}
    \caption{Ablation study of our model components on the LVIS benchmark. The results show the final AP after $K=2$ interactive turns. Removing either the intent state (IS) memory or the negative feedback mechanism causes a dramatic drop in performance, demonstrating that both are essential for effective ambiguity resolution. Our full model significantly outperforms all degraded variants.}
    \label{component}
\end{figure}

\begin{figure}
    \centering
    \includegraphics[width=0.92\linewidth]{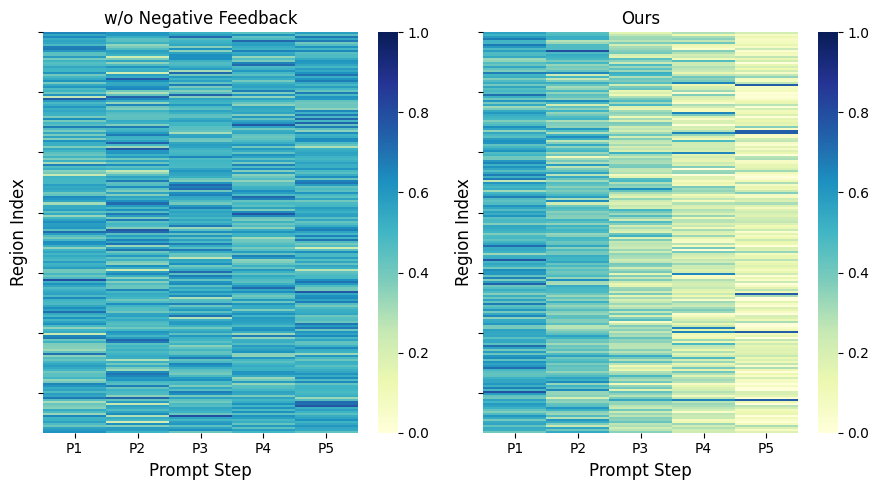}
    \caption{Visualization of the contrastive refinement process. This shows the normalized scores of 100 candidate regions (y-axis) over 5 interactive steps (x-axis) for an ambiguous query with multiple distractors. (Left) Without our contrastive mechanism \((\lambda=0)\), the model is confused; scores for many distractor regions stay high throughout the interaction. (Right) Our full model uses negative feedback to suppress the scores of irrelevant regions (light yellow) while specifying the score of the true target (dark blue).}
    \label{heatmap}
\end{figure}

\vspace{-10pt}

\subsection{Ablation Studies}
We conduct a series of ablation studies to validate the key components of our framework: the intent state and our contrastive ranking function. For this, we evaluate several variants of our model on the LVIS benchmark, reporting the AP score achieved after a maximum of \(K=2\) interactive turns in Figure \ref{component}. We compare the performance of our model with its variants. i) Our proposed model, which uses the intent state and the full contrastive score, \(\max(pos)-\lambda\cdot \max(neg)\). ii) w/o negative feedback, where we set \(\lambda=0\), then the score becomes a similarity search to the positive anchors $\max(pos)$. iii) w/ averaging score, where we replace our $\max$ operators with an averaging of exemplars, then the score becomes \(cos(r_j, \text{avg}(Z_{pos}))-\lambda\cdot cos(r_j, \text{avg}(Z_{neg}))\). iv) w/o intent state (stateless), that at each turn $t$, the intent state is reset and only contains the initial  prompt and the most recent user feedback. The results show that our full model significantly outperforms all versions. The most performance degradation occurs in the stateless variant (a drop of -10.8 AP), proving that the intent state is the most critical component for resolving complex ambiguity. Disabling negative feedback also weaken the model's performance (a drop of -5.9 AP), confirming that our contrastive learning from rejection is essential for disambiguation. Finally, our max-operator for ranking function outperforms the averaging strategy. 
Additionally, we compare our full model against a key ablation, which is disabling the negative feedback, and the result is presented in Figure \ref{heatmap}. For this experiment, we select a an example image from LVIS-Ambiguous benchmark with many similar objects. Then start with an initial text prompt (P1); for steps P2 through P5, simulate a user providing negative feedback on the current highest-scoring incorrect object. At each of the 5 steps, record the scores for all 100 candidate regions for both our model and its baseline (w/o negative feedback). As can be seen, our baseline model identifies many plausible regions at the start (P1), but because it cannot learn from negative feedback, the scores for the wrong objects (distractors) remain high throughout the interaction (P2-P5). However, our full model, at the start (P1), it is also confused and gives high scores to many regions. But as the user provides negative feedback in subsequent steps (P2, P3, P4), we can see the model learning, and the scores for the incorrect regions are suppressed and the correct target is identified.

\begin{table}
\centering
\begin{subtable}[t]{0.22\textwidth}
\centering
\begin{tabular}{c|c} 
\toprule
m &  AP (r/c/f)  \\ \midrule
25 &  34.5 (25.0 / 34.2 / 37.8)  \\
50 &  34.9 (25.7 / 34.3 / 38.0)    \\ \rowcolor{lightgray!35}
100 & 35.4 (25.6 / 34.7 / 38.2)    \\
200 & 34.9 (24.3 / 34.0 / 38.1)  \\ \bottomrule
\end{tabular}
\caption{Number of region candidate \\ (optimal: $m=100$)}
\label{tab:region_candidates}
\end{subtable}
\hfill
\begin{subtable}[t]{0.22\textwidth}
\centering
\begin{tabular}{c|c} 
\toprule
$\lambda$ & AP (r/c/f)  \\ \midrule
0.1  & 33.7 (24.2 / 33.6 / 37.5)  \\  
0.5  & 35.6 (25.0 / 34.5 / 37.9)    \\\rowcolor{lightgray!35}
1.0  & 35.4 (25.6 / 34.7 / 38.2) \\ 
1.5  & 35.1 (24.9 / 34.2 / 38.0)  \\
\bottomrule
\end{tabular}
\caption{Contrastive loss weight \\ (optimal: $\lambda=1$)}
\label{tab:contrastive_loss}
\end{subtable}
\caption{Hyperparameter sensitivity on LVIS/ResNet-50.}
\label{param}
\end{table}


\subsubsection{Hyperparameter sensitivity}
Table~\ref{param} analyzes the effect of two key hyperparameters: the number of region candidates, $m$ and the weight of the contrastive loss, $\lambda$. When $m=25$, the model achieves the lowest AP on rare categories (25.0), due to insufficient proposal coverage. Increasing to $m=100$ provides the best result, however, setting $m=200$ degrades performance on rare categories by 1.3\%, likely due to the inclusion of low-quality or noisy regions. For contrastive loss, $\lambda=0.5$ improves overall AP (35.6), but have slightly weaker performance on rare categories. We find that $\lambda \in [0.5, 1.0]$ provides balanced results, we set it to $1$. In terms of efficiency, a single interactive turn introduces only $\approx$29 ms of additional computation on an NVIDIA RTX~3090 GPU, which corresponds to less than 15\% of the total inference time. This confirms that the proposed feedback mechanism offers substantial accuracy gains with minimal computational overhead.




\section{Conclusion and Future work} 
In this work, a new framework is proposed for interactive object localization that can overcome a key limitation of current open-vocabulary detectors, which often fail on ambiguous queries involving multiple similar objects. We addressed this by reframing object retrieval as a stateful learning process, allowing the model to iteratively refine the user intent. The effectiveness of our model is driven by its intent state module, which accumulates positive and negative feedback, and a contrastive alignment function that leverages this memory to disambiguate between visually similar objects. Our experiments demonstrate that even a single corrective feedback significantly improves retrieval accuracy on standard open-world benchmarks, and especially on the challenging LVIS-Ambiguous subset. 
However, our model remains limited by its reliance on the initial set of candidate regions. If the detector fails to generate a bounding box for the true target (e.g., when the object is too small or heavily occluded), the interactive refinement process cannot recover it. In future work, we plan to explore mechanisms for updating or refining the candidate proposals based on user feedback. 


\bibliographystyle{ACM-Reference-Format} 
\bibliography{sample}


\end{document}